\documentclass[10pt,twocolumn,letterpaper]{article}

\usepackage{authblk}
\usepackage{cvpr}              

\usepackage{graphicx}
\usepackage{amsmath}
\usepackage{amssymb}
\usepackage{booktabs}

\usepackage{url}

\usepackage[accsupp]{axessibility}

\usepackage{mathrsfs}
\usepackage[mathscr]{eucal}
\usepackage{multirow}
\usepackage{diagbox} 
\usepackage{verbatim}
\usepackage{algorithm}
\usepackage{algorithmic}
\usepackage{wrapfig} 
\usepackage{pifont}
\usepackage{ctable}
\usepackage{marvosym}

\newtheorem{thm}{Theorem}[section]

\newtheorem{mylem}[thm]{Lemma}

\newtheorem{myprop}[thm]{Proposition}

\newtheorem{innercustompro}{Proof}
\newenvironment{custompro}[1]
  {\renewcommand\theinnercustompro{#1}\innercustompro}
  {\endinnercustompro}
\usepackage[pagebackref,breaklinks,colorlinks]{hyperref}

\definecolor{green(pigment)}{rgb}{0.1607, 0.3843, 0.0941}
\definecolor{blue(pigment)}{rgb}{0., 0.1484, 0.6992}
\definecolor{blue(back)}{rgb}{0.9058, 0.9019, 0.9725}
\definecolor{orange(pigment)}{rgb}{0.6, 0.298, 0.}
\hypersetup{colorlinks,linkcolor=blue(pigment),urlcolor=black}

\renewcommand{\vec}[1]{\mathbf{#1}}

\newcommand{\D}{\mathcal{D}}
\newcommand{\U}{\vec{u}}
\newcommand{\w}{\vec{w}}
\newcommand{\ww}{\vec{w}}
\newcommand{\uu}{\vec{u}}

\newcommand{\s}{\vec{s}}
\newcommand{\St}{\mathcal{S}}

\newcommand{\W}{\vec{W}}

\newcommand{\KL}{\mathrm{KL}}

\newcommand{\E}{\mathbb{E}}

\newcommand{\WW}{\mathbb{W}}

\newcommand{\Lc}{\mathcal{L}}

\newcommand{\ul}{\vec{u}}

\newcommand{\PP}{\mathrm{\mathbb{P}}}

\newcommand{\commentout}[1]{}

\usepackage{xcolor}

\definecolor{bluee}{rgb}{0., 0.1484, 0.6992}

\usepackage[capitalize]{cleveref}
\crefname{section}{Sec.}{Secs.}
\Crefname{section}{Section}{Sections}
\Crefname{table}{Table}{Tables}
\crefname{table}{Tab.}{Tabs.}

\def\cvprPaperID{6754} 
\def\confName{CVPR}
\def\confYear{2023}

\begin{document}

\title{Randomized Adversarial Training via Taylor Expansion}

\author[1,2]{Gaojie Jin}
\author[2\Letter]{Xinping Yi}
\author[2]{Dengyu Wu}
\author[3]{Ronghui Mu}
\author[2\Letter]{Xiaowei Huang}
\affil{State Key Laboratory of Computer Science, Institute of Software, CAS, Beijing, China}
\affil[2]{University of Liverpool, Liverpool, UK}
\affil[3]{Lancaster University, Lancaster, UK \authorcr
\{g.jin3, xinping.yi, xiaowei.huang\}@liverpool.ac.uk \authorcr
\textsuperscript{\Letter} Corresponding Author}
\renewcommand*{\Authands}{, }

\maketitle

\begin{abstract}
In recent years, there has been an explosion of research into developing more robust deep neural networks against adversarial examples. 
Adversarial training appears as one of the most successful methods. 
To deal with both the robustness against adversarial examples and the accuracy over clean examples, many works  
develop enhanced adversarial training methods to  
achieve various trade-offs between them \cite{engstrom2018evaluating,kannan2018adversarial,zhang2019theoretically}.  
Leveraging over the studies \cite{DBLP:conf/uai/IzmailovPGVW18,chen2020robust} that smoothed update on weights during training may help find flat minima and improve generalization, we suggest reconciling the robustness-accuracy trade-off from another perspective,   
i.e., by adding random noise into deterministic weights.
The randomized weights  
enable our design of a novel adversarial training method via Taylor expansion of a small Gaussian noise, and we show that the new adversarial training method can flatten loss landscape and find flat minima.  
With PGD, CW, and Auto Attacks, an extensive set of experiments demonstrate that our method 
enhances the state-of-the-art adversarial training methods, boosting both robustness and clean accuracy.
The code is available at \url{https://github.com/Alexkael/Randomized-Adversarial-Training}.
\end{abstract}

\vspace{-4mm}
\section{Introduction}
\begin{figure*}[t!]
\includegraphics[width=1
\textwidth]{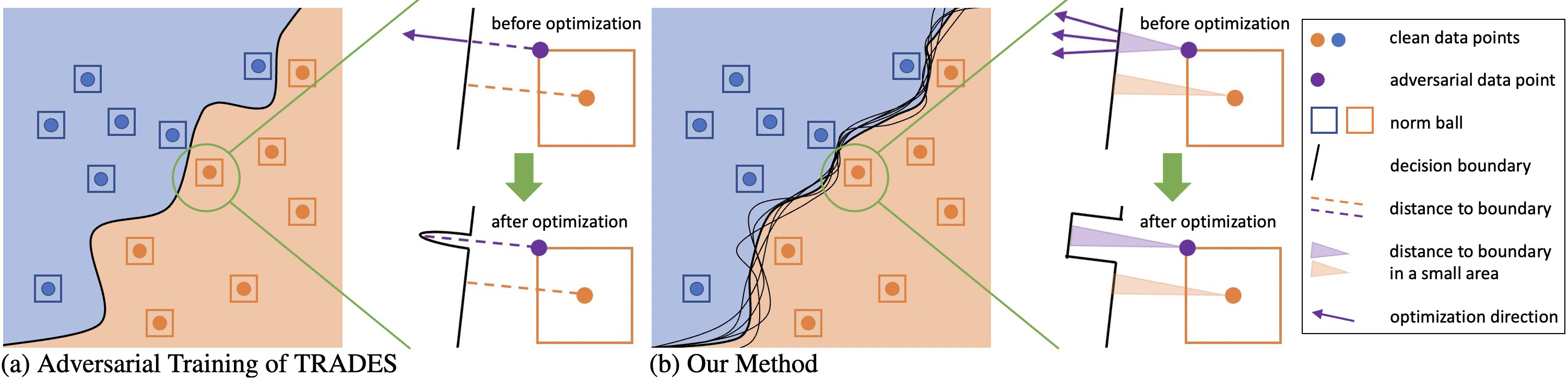}
\centering
\vspace{-6mm}
\caption{A conceptual illustration of decision boundaries learned via \textbf{(a)} adversarial training of TRADES and \textbf{(b)} our method. 
\textbf{(a)} shows that TRADES considers a deterministic model and optimizes the distance of adversarial data and boundary only through one direction. 
Our method in \textbf{(b)} takes into account randomized weights (perturbed boundaries) and optimizes the distance of adversarial data and boundary via multi directions in a small area. The boundary learned by our method can be smoother and more robust against new data.}
\vspace{-5mm}
\label{fig:1}
\end{figure*}

\vspace{-2mm}
Trade-off between adversarial robustness and clean accuracy has recently been intensively studied~\cite{DBLP:conf/nips/SchmidtSTTM18,DBLP:conf/eccv/SuZCYCG18,zhang2019theoretically} 
and demonstrated to exist
\cite{DBLP:conf/iclr/TsiprasSETM19,DBLP:conf/icml/RaghunathanXYDL20,DBLP:conf/colt/JavanmardSH20,DBLP:journals/corr/abs-2103-09947}. 
Many different techniques have been 
developed to 
alleviate the loss of clean accuracy when improving robustness, 
including data augmentation~\cite{DBLP:conf/nips/AlayracUHFSK19,DBLP:conf/nips/CarmonRSDL19,DBLP:conf/icml/HendrycksLM19}, early-stopping~\cite{DBLP:conf/icml/RiceWK20,DBLP:conf/icml/ZhangXH0CSK20}, instance reweighting~\cite{DBLP:journals/corr/abs-1910-08051,DBLP:conf/iclr/ZhangZ00SK21}, and 
various 
adversarial training methods~\cite{DBLP:conf/nips/WuX020,DBLP:conf/cvpr/LeeLY20,DBLP:conf/iccv/Cui0WJ21,DBLP:journals/corr/abs-2203-06020,zhang2019theoretically,mu4251640sparse}. 
Adversarial training is believed to be the most effective defense method against adversarial attacks, 
and is usually formulated as a minimax optimization problem 
where the network weights are assumed deterministic in each alternating iteration. 
Given the fact that both clean and adversarial examples are drawn from unknown distributions which interact with one another through the network weights, it is 
reasonable to relax the assumption that neural networks 
are \emph{simply} deterministic models where weights are  scalar values. 
This paper is based on a view, illustrated in \cref{fig:1}, that 
 randomized models 
 enable the training optimization  to consider multiple directions within a small area and may 
 achieve smoothed weights update -- in a way different from checkpoint averaging \cite{DBLP:conf/uai/IzmailovPGVW18,chen2020robust} -- and obtain robust models against new clean/adversarial examples.

Building upon the above view, 
we find a way, drastically different from most existing studies in adversarial training, to balance robustness and clean accuracy, that is, \emph{embedding neural network weights with random noise}. 
Whilst the randomized weights framework is not new in statistical learning theory, where it has been 
used in many previous works for e.g., generalization analysis~\cite{DBLP:conf/uai/DziugaiteR17,DBLP:conf/nips/NeyshaburBMS17,xu2017information}, 
we hope to advance the empirical understanding of the robustness-accuracy trade-off problem in adversarial training by leveraging the rich tool sets in statistical learning.
Remarkably, it turns out 
adversarial training with the optimization over randomized weights can 
improve the state-of-the-art adversarial training methods over both the adversarial robustness and clean accuracy.

By modeling 
weights as randomized variables with an artificially injected weight perturbation, we start with an empirical analysis of flatness of loss landscape in \cref{sec:motivation}.
We show that our method can 
flatten loss landscape and find flatter minima in adversarial training, which is 
generally regarded as an indicator 
of good generalization ability.
After the 
flatness analysis, we show \emph{how to optimize with randomized weights during adversarial training} in Sec.~\ref{sec:method}.
A novel adversarial training method based on TRADES \cite{zhang2019theoretically} is proposed to reconcile adversarial robustness with clean accuracy by closing the gap between clean latent space and adversarial latent space over randomized weights. 
Specifically, we utilize Taylor series to expand the objective function over weights, in such a way that we can deconstruct the function into Taylor series (e.g., zeroth term, first term, second term, etc). 
From an algorithmic viewpoint, these Taylor terms can thus replace the objective function effectively and time-efficiently.
As \cref{fig:1} shows, since our method takes randomized models into consideration during training, the learned boundary is smoother and the learned model is more robust in a small perturbed area.

We {validate the effectiveness of our optimization method} with the first 
and the second derivative terms of Taylor series.
In consideration of training complexity and efficiency, we omit the third and higher derivative terms. Through an extensive set of experiments on a wide range of datasets (CIFAR-10~\cite{krizhevsky2009learning}, CIFAR-100, SVHN~\cite{netzer2011reading}) and model architectures (ResNet~\cite{he2016deep}, WideResNet~\cite{DBLP:conf/bmvc/ZagoruykoK16}, VGG~\cite{DBLP:journals/corr/SimonyanZ14a}, MobileNetV2~\cite{sandler2018mobilenetv2}), we find that \emph{our method can further enhance the state-of-the-art  adversarial training methods on both adversarial robustness and clean accuracy, consistently across the datasets and the model architectures}.
Overall, this paper makes the following contributions: 
\vspace{-2mm}
\begin{itemize}
    \setlength{\itemsep}{0pt}
    \setlength{\parsep}{0pt}
    \setlength{\parskip}{0pt}
    \item We conduct a pilot study of the trade-off between adversarial robustness and clean accuracy with randomized weights, and offer a new insight on the smoothed weights update and the flat minima during adversarial training (\cref{sec:motivation}).
    \item We propose a novel adversarial training method under a randomized model to smooth the weights. The key enabler is the Taylor series expansion (in Sec.~\ref{sec:method}) of the robustness loss function over randomized weights (deterministic weights with random noise), so that the optimization can be simultaneously done over the zeroth, first, and second orders of Taylor series. 
    In doing so, the proposed method can effectively enhance adversarial robustness without a significant compromise on clean accuracy.
    \item An extensive set of empirical results are provided to demonstrate that our method can improve both robustness and clean accuracy consistently across different datasets and different network architectures (Sec.~\ref{sec:experiment}).
\end{itemize}

\section{Preliminaries}
\label{sec:preliminaries}
\textbf{Basic Notation.} Consider the classification task with the training set $\St=\{\s_1,...,\s_m \}$ where $m$ samples are drawn from the data distribution $\D$. 
For notational convenience, we omit the label $y$ of sample $\s$. 
The adversarial example $\s'$ is generally not in the natural dataset $\St$, such that $||\s'-\s||_p\le \epsilon$ where $||\cdot||_p$ is by default the $\ell_p$-norm.
Let $f_{\w}(\cdot)$ be a learning model parameterized by $\w$, where $\w$ is the vectorization of weight matrix $\W$.
$\Lc(f_\w(\s), y)$ is the cross-entropy loss between $f_\w(\s)$ and $y$ with normalization. 
Define $\Lc(f_\w(\St), \mathcal{Y}):=\E_{\s\in \St}[\Lc(f_\w(\s), y)]$. 

\textbf{Vanilla Adversarial Training.} Adversarial training can be formulated as a minimax optimization problem~\cite{DBLP:conf/iclr/MadryMSTV18}
\vspace{-2mm}
\begin{equation}
\label{eq:15}
    \min_{\w}\Big\{\mathop{\E}\limits_{\s\sim \D} \Big[\max_{\s':||\s'-\s||_p\le \epsilon} \Lc(f_\w(\s'),y)\Big]\Big\},
\vspace{-2mm}
\end{equation}
where $\s'$ is an adversarial example causing the largest loss within an $\epsilon$-ball centered at a clean example $\s$ with respect to a norm distance. 
Adversarial training aims to learn a model with the loss of adversarial examples minimized, without caring about the loss of the clean examples.


\begin{figure*}[t!]
\includegraphics[width=0.8
\textwidth]{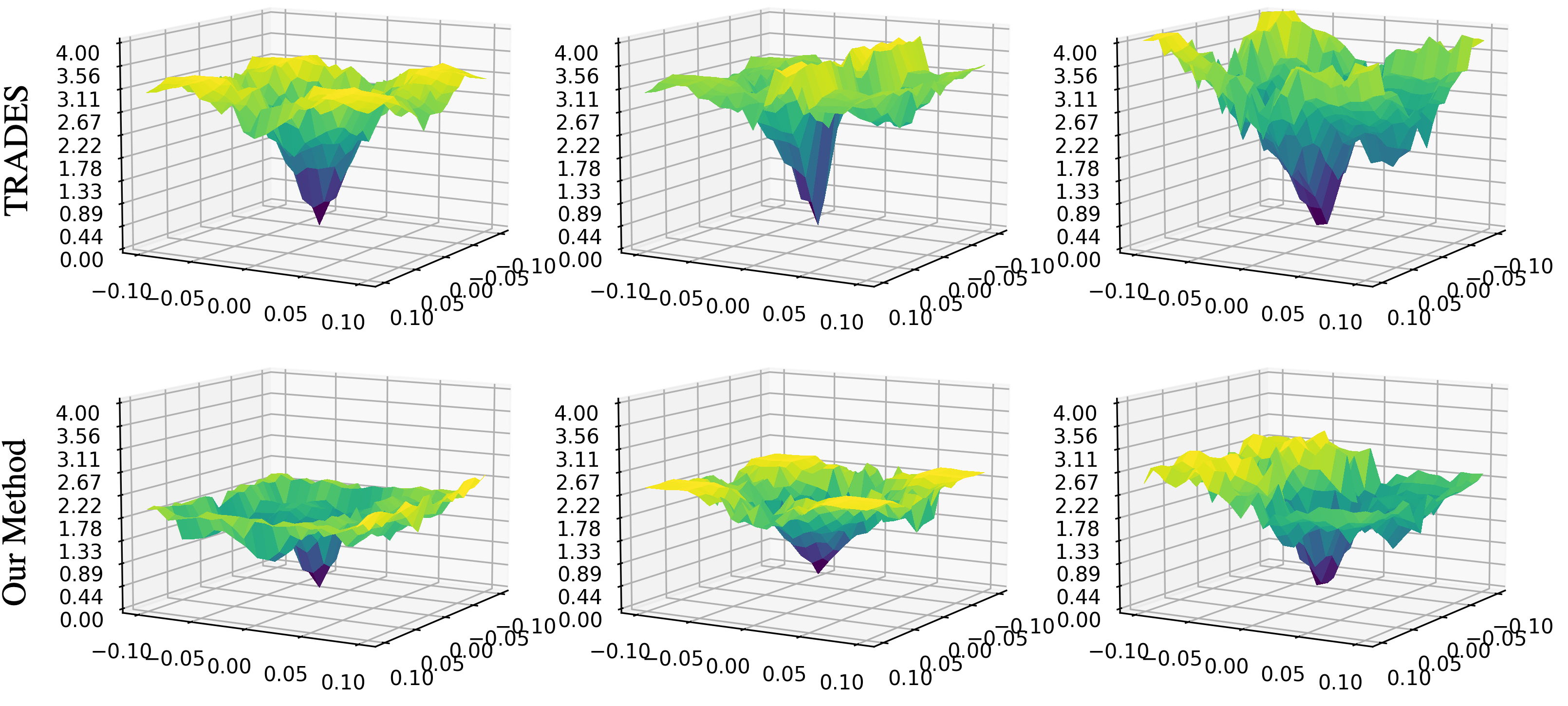}
\centering
\vspace{-4mm}
\caption{Comparison of loss landscapes of TRADES trained model (the first row) and TRADES+$1_{st}$+$2_{nd}$ (our method, the second row) trained model.
Loss plots in each column are generated from the same original image randomly chosen from the CIFAR-10 test dataset.
Following the settings in \cite{DBLP:journals/corr/abs-1807-10272}, the $z$ axis represents the loss, the $x$ and $y$ axes represent the magnitude of the perturbation added in the directions of $\vec{sign}\nabla_{\s}f(\s)$ and Rademacher($0.5$) respectively. 
We provide more empirical results in \cref{appendix:a}.}
\vspace{-4mm}
\label{fig:2}
\end{figure*}

\textbf{TRADES-like Adversarial Training.} Advanced adversarial training methods, such as TRADES \cite{zhang2019theoretically}, strike a trade-off between clean accuracy and adversarial robustness, by optimizing the following problem
\begin{small}
\begin{equation}\nonumber
\label{eq:16}
    \min_{\w}\Big\{\mathop{\E}\limits_{\s\sim \D} \Big[ \Lc(f_\w(\s),y) + \max_{\s':||\s'-\s||_p\le \epsilon}\Lc(f_\w(\s),f_\w(\s'))/\lambda\Big]\Big\},
\end{equation}
\end{small}where the first term contributes to clean accuracy, and the second term with hyperparameter $\lambda$ can be seen as a regularization for adversarial robustness that balances the outputs of clean and adversarial examples.

Nevertheless, each adversarial example $\s'$ is generated from a clean example $\s$ based on a specific weight $\w$, and the weight $\w$ is obtained by optimizing both clean and adversarial data with SGD. 
This is a chick-and-egg problem. A not-very-well-trained model $\w$ may generate some misleading adversarial data samples, which in turn drive the training process away from the optimum.

Inspired by the idea of model smoothing, we believe that the introduction of noise term into the training process may smooth the update of weights and thus robustify the models. To elaborate on this, we introduce our method from the perspective of flatness of loss landscape in Sec.~\ref{sec:motivation}.
This is followed by a formalization in Sec.~\ref{sec:method}, where a noise term is injected to network weight $\w$ during adversarial training to transform  deterministic weights into randomized weights. 

\vspace{-1mm}
\section{Motivation}
\vspace{-1mm}
\label{sec:motivation}

Flatness is commonly thought to be a metric of standard generalization: the loss surface at the final learnt weights for well-generalizing models is generally ``flat" \cite{DBLP:conf/uai/DziugaiteR17,DBLP:conf/nips/NeyshaburBMS17,DBLP:conf/iclr/FarniaZT19,DBLP:conf/iclr/JiangNMKB20}.
Moreover, \cite{DBLP:journals/corr/abs-2004-05884} believed that a flatter adversarial loss landscape reduces the generalization gap in robustness.
This aligns with \cite{DBLP:conf/nips/HeinA17}, where the authors 
demonstrated that the local Lipschitz constant can be used to explicitly quantify the robustness of machine learning models.
Many empirical defensive approaches, such as hessian/curvature-based regularization \cite{DBLP:conf/cvpr/Moosavi-Dezfooli19}, gradient magnitude penalty \cite{DBLP:conf/iccv/WangZ19}, smoothening with random noise \cite{DBLP:conf/eccv/LiuCZH18}, or entropy regularization \cite{DBLP:journals/frai/JagatapJCGH21}, have echoed the flatness performance of a robust model.
However, all the above approaches require significant computational or memory resources, and many of them, such as hessian/curvature-based solutions, may suffer from standard accuracy decreases \cite{DBLP:conf/iclr/GuptaSD20}.

Stochastic weight averaging (SWA) \cite{DBLP:conf/uai/IzmailovPGVW18} is known to find solutions that are far flatter than SGD, is incredibly simple to implement, enhances standard generalization, and has almost no computational overhead.
It was developed to guarantee weight smoothness by simply averaging various checkpoints throughout the training trajectory.
SWA has been applied to semi-supervised learning, Bayesian inference, and low-precision training.
Especially, \cite{chen2020robust} used SWA for the first time in adversarial training in order to smooth the weights and locate flatter minima, and showed that it enhances the adversarial generalization.

Different from SWA, we propose a novel method (with details provided in \cref{{sec:method}}) via Taylor expansion of a small Gaussian noise to smooth the update of weights. This method regards the TRADES \cite{zhang2019theoretically}, a state-of-the-art adversarial training method, as a special case with only zero-th term, and is able to locate flatter minima. To demonstrate the latter, we visualize the loss landscape with respect to both input and weight spaces.
As shown in \cref{fig:2}, compared to the TRADES, our method significantly flattens the rugged landscape.
In addition, it is worth noting that our empirical results (in \cref{sec:experiment}) demonstrate that our method can further improve not only clean accuracy but also adversarial accuracy compared with other SOTA (includes SWA from \cite{chen2020robust}).

\begin{algorithm*}[t]
\caption{Adversarial training with randomized weights}
\label{alg}
\hspace*{0.1in}\textbf{Input:} minibatch $\{\s_i\}_{i=1}^n$, network architecture parametrized by $\w$, learning rate $\eta_l$, step size $\eta_s$, hyper-parameters $\eta$, $\lambda$,\\
\hspace*{0.2in} zero mean Gaussian $\mathbf{u}$, number of iterations $K$ in inner optimization.\\
\hspace*{0.1in}\textbf{Output:} Robust network $f_\w$\\
\hspace*{0.1in}Randomly initialize network $f_\w$, or initialize network with pre-trained configuration\\
\hspace*{0.1in}\textbf{repeat} \\
\hspace*{0.2in}$\blacktriangleright$ Generate adversarial examples: \\
\hspace*{0.3in} Sample $u$ from $\uu$ \\
\hspace*{0.3in} \textbf{for} $i=1$ to $n$ \textbf{do} \\
\hspace*{0.5in} $\s'_i\gets \s_i+0.001\cdot \mathcal{N}(\mathbf{0},\mathbf{I})$ \\
\hspace*{0.5in} \textbf{for} $k=1$ to $K$ \textbf{do} \\
\hspace*{0.7in} $\s'_i\gets \Pi(\eta_s\vec{sign}(\nabla_{\s'_i}\Lc(f_{\w+u}(\s_i),f_{\w+u}(\s'_i)))+\s'_i)$, subject to $||\s_i-\s_i'||_p\le\epsilon$, where $\Pi$ is the projection\\
\hspace*{0.7in} operator.\\
\hspace*{0.5in} \textbf{end for} \\
\hspace*{0.3in} \textbf{end for} \\
\hspace*{0.2in}$\blacktriangleright$ Optimization: $\w\gets \w-\eta_l \frac{1}{n}\sum_{i=1}^n \nabla_\w \Big[\Lc(f_\w(\s_i),y_i)$\\
\hspace*{1.22in} {\color{blue(pigment)}(zeroth term)}\quad\quad\quad\quad$+{\color{blue(pigment)}\Lc(g_{\s_i}(\w),g_{\s'_i}(\w))/\lambda}$\\
\hspace*{1.22in} {\color{orange(pigment)}(first term)}\quad\quad\quad\;\;\;\;\;\; $+{\color{orange(pigment)}\eta \E_{\mathbf{u}}\big(\Lc(g'_{\s_i}(\w)^T\mathbf{u},g'_{\s'_i}(\w)^T \mathbf{u})\big)}$\\
\hspace*{1.22in} {\color{green(pigment)}(second term)}\quad\quad\quad\,\,\, $+{\color{green(pigment)}\frac{\eta}{2}\E_{\mathbf{u}}\big(\Lc(\mathbf{u}^Tg''_{\s_i}(\w)\mathbf{u},\mathbf{u}^Tg''_{\s'_i}(\w)\mathbf{u})\big)}\Big]$,\\
\hspace*{0.3in} where various optimization methods can be applied: {\color{blue(pigment)}zeroth} term optimization (TRADES), {\color{blue(pigment)}zeroth} + {\color{orange(pigment)}first} terms opti-\\
\hspace*{0.3in} mization and {\color{blue(pigment)}zeroth} + {\color{orange(pigment)}first} + {\color{green(pigment)}second} terms optimization. \\
\hspace*{0.1in}\textbf{until} training converged
\end{algorithm*}

\vspace{-1mm}
\section{Methodology}
\vspace{-1mm}
\label{sec:method}

Given a deterministic model of weight $\w$, we consider an additive small Gaussian noise $\uu$ to smooth $\w$.
Based on TRADES~\cite{zhang2019theoretically}, we propose that the robust optimization problem can be further enhanced by
\vspace{-2mm}
\begin{equation}
\begin{aligned}
\label{eq:10}
    \min_\w \; &\E_{\uu}\Big[ \E_\s (\Lc(f_{\w+\uu}(\s),y))\\ 
    &+ \E_\s \max_{\s': ||\s'-\s||_p\le \epsilon}\KL(f_{\w+\uu}(\s)||f_{\w+\uu}(\s'))/\lambda \Big], 
\end{aligned}
\vspace{-2mm}
\end{equation}
where the first term contributes to clean accuracy, and the second term with hyperparameter $\lambda$ can be seen as a regularization for adversarial robustness with weights $\w$ and a small Gaussian noise $\uu$.

Following \cite{DBLP:conf/iclr/JiangNMKB20,DBLP:conf/nips/DziugaiteDNRCWM20}, we let the Gaussian $\uu$ be small enough to maintain the generalization performance on clean data.
Then, we can extend \cref{eq:10} by replacing $\KL(\cdot ||\cdot)$ with a multi-class calibrated loss $\Lc(\cdot,\cdot)$ \cite{zhang2019theoretically}:
\begin{equation}
\begin{aligned}
    \min_\w \; &\E_{\uu}\Big[ \E_\s (\Lc(f_{\w}(\s),y))\\
    &+ \E_\s \max_{\s': ||\s'-\s||_p\le \epsilon}\Lc(f_{\w+\uu}(\s),f_{\w+\uu}(\s'))/\lambda \Big]. 
\end{aligned}
\end{equation}

As the minimax optimization is entangled with expectation of randomized weights, it is challenging to solve such optimization problem directly. Instead, we propose to solve this problem in an alternating manner between the inner maximization and outer minimization. 

For the inner maximization, with a given model weight $\w$, we solve it by adopting the commonly used gradient-based approach (e.g., PGD) to generate the adversarial perturbation. That is, at the $t$-th iteration, the adversarial perturbation is produced iteratively by letting
\begin{equation}
\s'_{t+1}\gets \Pi\Big(\s'_t + \eta_s\vec{sign}\big(\nabla_{\s'_t}\Lc(f_{\w+u}(\s),f_{\w+u}(\s'_t))\big)\Big)
\end{equation}
until the maximum allowed iterations are reached, where $\Pi$ is the projection operator, $\eta_s$ is the step size and $u$ is a sample of $\uu$.
To reduce complexity, we only use one sample $u$ to generate adversarial example for each minibatch.
Nevertheless, sufficient samples of $\uu$ can still be produced in one epoch as there are many minibatches.
As Fig.~\ref{fig:1} shows, the adversarial example is generated by one model (boundary), then the optimization with randomized weights in multiple directions can be executed through the following minimization method.

For the outer minimization, when the optimal perturbed sample $\s'$ is generated by the inner maximization, we solve the following problem
\begin{equation} \label{eq:outer}
    \min_{\w} \; \E_\s \Big[\Lc(f_{\w}(\s),y) + \E_{\uu} \Lc(f_{\w+\uu}(\s),f_{\w+\uu}(\s'))/\lambda \Big].
\end{equation}
This optimization problem is challenging due to the entanglement between the deterministic weight $\w$ and the random perturbation $\uu$, which makes gradient updating much involved.

To resolve this issue, we decompose the two components in such a way that gradient updating and model averaging can be done separately.
Let $f_{\w+\uu}(\s)=g_\s(\w+\uu)$, by Taylor expansion of $g_\s(\w+\uu)$ at $\w$, we have
\begin{equation}
\begin{aligned} 
    g_{\s}(\w+\mathbf{u})=g_{\s}(\w)&+g'_{\s}(\w)^T\mathbf{u}+\mathbf{u}^T\frac{g''_{\s}(\w)}{2!}\mathbf{u}\\
    &+{\mathcal{O}_{\s}}(||\mathbf{u}||^2),
\end{aligned}
\end{equation}
where $g'_{\s}(\w)=\frac{\partial g_{\s}(\w)}{\partial \w} \in \mathbb{R}^{d \times 1}$ and $g''_{\s}(\w)=\frac{\partial^2 g_{\s}(\w)}{\partial \w \partial \w} \in \mathbb{R}^{d \times d}$ are the first and second derivative of the function $g_{\s}(\w)$ versus the weight $\w$, respectively, and ${\mathcal{O}_{\s}}(||\mathbf{u}||^2)$ tends to be zero with respect to $||\mathbf{u}||^2$. By such approximation, the gradient computation will be solely done on the deterministic part $\w$, and model smoothing with averaged random variable $\mathbf{u}$ is done independently.

The intuition behind is the following. For the randomized model $\w+\mathbf{u}$ with a specific input $\s$, the Taylor approximation explicitly takes into account the deterministic model $g_{\s}(\w)$, the projection of its gradient onto the random direction $g'_{\s}(\w)^T\mathbf{u}$, and the projection of its curvature onto the subspace spanned by $\mathbf{u}^T\frac{g''_{\s}(\w)}{2!}\mathbf{u}$, capturing higher-order information of the loss function with respect to $\U$.

To further simplify the computation, we minimize an upper bound of Eq.~\eqref{eq:outer} instead.
Due to the local convexity of loss landscape, by Jensen's inequality, we conclude that
\begin{equation}
    \Lc\big(\sum_i x_i, \sum_i y_i \big) \le \sum_i \Lc(x_i, y_i).
\end{equation}
Therefore, the second term in Eq.~\eqref{eq:outer} can be upper bounded by
\vspace{-2mm}
\begin{equation}
\begin{aligned}
    &\E_{\mathbf{u}} \Lc(f_{\w+\uu}(\s),f_{\w+\uu}(\s'))\\ &\le \Lc(g_{\s}(\w),g_{\s'}(\w)) + \E_{\mathbf{u}}\big(\Lc(g'_{\s}(\w)^T\mathbf{u},  g'_{\s'}(\w)^T\mathbf{u})\big) \\
    &\quad +\frac{1}{2}\E_{\mathbf{u}}\big(\Lc(\mathbf{u}^Tg''_{\s}(\w)\mathbf{u},\mathbf{u}^Tg''_{\s'}(\w)\mathbf{u})\big), \label{eq:upper}
\end{aligned}
\end{equation}
where the terms on the right-hand side are referred to as the zeroth, first, and second order Taylor expressions.
Instead of minimizing \cref{eq:outer} directly, we minimize the upper bound in \cref{eq:upper}, which is easier to compute in practical models.
Details of the optimization are given in \Cref{appendix:d}.

The pseudo-code of the proposed adversarial training method is presented in \cref{alg}. Note that the zeroth term optimization in \cref{alg} is almost the same as TRADES~\cite{zhang2019theoretically}, thus we use TRADES and AWP-TRADES~\cite{DBLP:conf/nips/WuX020} as (zeroth term optimization) baselines in our experiments in Sec.~\ref{sec:experiment}, and verify how much first and second terms optimization can improve. 
Note that we perform both AWP and our method through applying randomized weight perturbations on AWP updated weights. 


For all of our experiments, we limit the noise variance of $\mathbf{u}$ to be small enough \cite{DBLP:conf/iclr/JiangNMKB20,DBLP:conf/nips/DziugaiteDNRCWM20} to maintain the training, e.g., $\sigma=0.01$. For the zeroth term hyper-parameter $1/\lambda$ in Algorithm~\ref{alg}, we set $1/\lambda=6$ for all of our experiments, which is a widely-used setting for TRADES~\cite{zhang2019theoretically,DBLP:conf/nips/WuX020,DBLP:journals/corr/abs-2202-10103}. The number of inner optimization iterations $K$ is set to $10$ as usual. 
To make our method more flexible, we set $\eta$ and $\frac{\eta}{2}$ as the first term and second term hyper-parameters, respectively, and analyze the sensitivity of $\eta$ in Sec.~\ref{subsec:5.1}.

The main novelty of our proposed method is two-fold:
\textbf{(1)} Instead of considering a single model with a deterministic weight $\w$, we consider a model ensemble with $\w+\uu$ to smooth the update of weights. 
By averaging these models during adversarial training, we come up with a robust model with smoothed classification boundary tolerant to potential adversarial weight perturbation. 
\textbf{(2)} When averaging over the ensemble of randomized models during training, we disentangle gradient from random weight perturbation so that the gradient updating can be computed efficiently. In particular, we apply Taylor expansion at the mean weight $\w$ (i.e., the deterministic component) and approximate the cross-entropy loss function with the zeroth, first, and second Taylor terms. In doing so, the deterministic and statistical components of $\w+\ul$ can be decomposed and then computed independently.

\section{Empirical results}
\label{sec:experiment}

In this section, we first discuss the hyper-parameter sensitivity of our method, and then evaluate the robustness on benchmark datasets against various white-box, black-box attacks and Auto Attack with $\ell_2$ and $\ell_\infty$ threat models.

\textbf{Adversarial Training Setting.} We train PreAct ResNet-18~\cite{he2016deep} for $\ell_\infty$ and $\ell_2$ threat models on CIFAR-10/100 \cite{krizhevsky2009learning} and SVHN \cite{netzer2011reading}. 
In addition, we also train  WideResNet-34-10~\cite{DBLP:conf/bmvc/ZagoruykoK16} for CIFAR-10/100, VGG16 and MobileNetV2 for CIFAR-10, with $\ell_\infty$ threat model.
We adopt the widely used adversarial training setting~\cite{DBLP:conf/icml/RiceWK20}: for the $\ell_\infty$ threat model, $\epsilon=8/255$ and step size $2/255$;
for the $\ell_2$ threat model, $\epsilon=128/255$ and step size $15/255$.
For normal adversarial training, the training examples are generated with 10 steps.
All models (except SVHN) are trained for $200$ epochs using SGD with momentum $0.9$, batchsize $128$, weight decay $5\times 10^{-4}$, and an initial learning rate of $0.1$ that is divided by $10$ at the 100th and 150th epochs.
Except for setting the starting learning rate to $0.01$ for SVHN, we utilize the same other settings.
Simple data augmentations are used, such as $32\times 32$ random crop with $4$-pixel padding and random horizontal flip. 
We report the highest robustness that ever achieved at different checkpoints for each dataset and report the clean accuracy on the model which gets the highest PGD-20 accuracy. 
We omit the standard deviations of 3 runs as they are very small ($<0.40\%$) and implement all models on NVIDIA A100.

\begin{table*}[t!]
\centering
\caption{First and Second Derivative terms optimization on CIFAR-10 with $\ell_\infty$ threat model for PreAct ResNet18. Classification accuracy (\%) on clean images and under PGD-20 attack, CW-20 attack and Auto Attack with different hyper-parameters $\eta=0.05, 0.1, 0.2, 0.3, 0.4$.
We highlight the best results in \textbf{bold}.}
\label{tab:1}
\vspace{-3mm}
\scalebox{1}{
\begin{tabular}{lccccclcccc}
\specialrule{.15em}{.075em}{.075em} 
\multicolumn{1}{c}{Method} & Clean & PGD-20 & CW-20 & AA & & \multicolumn{1}{c}{Method} & Clean & PGD-20 & CW-20 & AA          \\ \cline{1-5} \cline{7-11} 
1$_{st}$ (0.05) & 83.19 & 53.98 & 52.12 & 48.2 & &  1$_{st}$+2$_{nd}$ (0.05) & 83.25 & 54.07 & 51.93 & 48.4  \\
1$_{st}$ (0.1) & 82.86 & 54.29 & 52.24 & 49.3 & &  1$_{st}$+2$_{nd}$ (0.1) & 83.47 & 54.42 & 52.28 & 49.7  \\
1$_{st}$ (0.2) & 83.55 & 54.86 & \textbf{52.65} & 48.8 & &  1$_{st}$+2$_{nd}$ (0.2) & 84.13 & \textbf{54.91} & \textbf{52.53} & \textbf{50.3}  \\
1$_{st}$ (0.3) & \textbf{83.96} & \textbf{55.05} & 52.54 & \textbf{49.7} & &  1$_{st}$+2$_{nd}$ (0.3) & \textbf{84.27} & 54.36 & 51.80 & 48.9   \\
1$_{st}$ (0.4) & 83.93 & 54.65 & 52.23 & 48.6 & &  1$_{st}$+2$_{nd}$ (0.4) & 84.14 & 54.38 & 51.56 & 49.6  \\
\specialrule{.15em}{.075em}{.075em} 
\end{tabular}
}
\vspace{-3mm}
\end{table*}
\begin{table*}[t!]
\centering
\caption{First and Second Derivative terms optimization on CIFAR-10/CIFAR-100 with $\ell_\infty$ threat model for WideResNet, compared with current state-of-the-art. Classification accuracy (\%) on clean images and under PGD-20 attack, CW-20 attack ($\epsilon=0.031$) and Auto Attack ($\epsilon=8/255$). The results of our methods are in \textbf{bold}. Note that $^*$ is under PGD-40 attack and $^{**}$ is under PGD-10 attack.}
\label{tab:2}
\vspace{-3mm}
\scalebox{1}{
\begin{tabular}{cclccccc}
\specialrule{.15em}{.075em}{.075em} 
Dataset & & \multicolumn{1}{c}{Method} & Architecture & Clean & PGD-20 & CW-20 & AA \\ \hline
\multirow{12}{*}{\shortstack{CIFAR-10\\$\ell_\infty$}} & & Lee et al. (2020) \cite{DBLP:conf/cvpr/LeeLY20} & WRN-34-10 & 92.56 & 59.75 & 54.53 & 39.70 \\
& & Wang et al. (2020) \cite{DBLP:conf/iclr/0001ZY0MG20} & WRN-34-10 & 83.51 & 58.31 & 54.33 & 51.10 \\
& & Rice et al. (2020) \cite{DBLP:conf/icml/RiceWK20} & WRN-34-20 & 85.34 & - & - & 53.42 \\
& & Zhang et al. (2020) \cite{DBLP:conf/icml/ZhangXH0CSK20} & WRN-34-10 & 84.52 & - & - & 53.51 \\
& & Pang et al. (2021) \cite{DBLP:conf/iclr/PangYDSZ21} & WRN-34-20 & 86.43 & 57.91$^{**}$ & - & 54.39 \\
& & Jin et al. (2022) \cite{DBLP:journals/corr/abs-2203-06020} & WRN-34-20 & 86.01 & 61.12 & 57.93 & 55.90 \\
& & Gowal et al. (2020) \cite{DBLP:journals/corr/abs-2010-03593} & WRN-70-16 & 85.29 & 58.22$^*$ & - & 57.20 \\
\cline{3-8}
& & Zhang et al. (2019) \cite{zhang2019theoretically} (0$_{th}$) & WRN-34-10 & 84.65 & 56.68 & 54.49 & 53.0 \\
& & \;\; + \textbf{Ours (1$_{st}$)} & WRN-34-10 & \textbf{85.51} & \textbf{58.34} & \textbf{56.06} & \textbf{54.0} \\
& & \;\; + \textbf{Ours (1$_{st}$+2$_{nd}$)} & WRN-34-10 & \textbf{85.98} & \textbf{58.47} & \textbf{56.13} & \textbf{54.2} \\
\cline{3-8}
& & Wu et al. (2020) \cite{DBLP:conf/nips/WuX020} (0$_{th}$) & WRN-34-10 & 85.17 & 59.64 & 57.33 & 56.2 \\
& & \;\; + \textbf{Ours (1$_{st}$)} & WRN-34-10 & \textbf{86.10} & \textbf{61.47} & \textbf{58.09} & \textbf{57.1} \\
& & \;\; + \textbf{Ours (1$_{st}$+2$_{nd}$)} & WRN-34-10 & \textbf{86.12} & \textbf{61.45} & \textbf{58.22} & \textbf{57.4} \\
\hline
\multirow{8}{*}{\shortstack{CIFAR-100\\$\ell_\infty$}} & & Cui et al. (2021) \cite{DBLP:conf/iccv/Cui0WJ21} & WRN-34-10 & 60.43 & 35.50 & 31.50 & 29.34 \\
& & Gowal et al. (2020) \cite{DBLP:journals/corr/abs-2010-03593} & WRN-70-16 & 60.86 & 31.47$^*$ & - & 30.03 \\
\cline{3-8}
& & Zhang et al. (2019) \cite{zhang2019theoretically} (0$_{th}$) & WRN-34-10 & 60.22 & 32.11 & 28.93 & 26.9  \\
& & \;\; + \textbf{Ours (1$_{st}$)} & WRN-34-10 & \textbf{63.01} & \textbf{33.26} & \textbf{29.44} & \textbf{28.1} \\
& & \;\; + \textbf{Ours (1$_{st}$+2$_{nd}$)} & WRN-34-10 & \textbf{62.93} & \textbf{33.36} & \textbf{29.61} & \textbf{27.9} \\
\cline{3-8}
& & Wu et al. (2020) \cite{DBLP:conf/nips/WuX020} (0$_{th}$) & WRN-34-10 & 60.38 & 34.09 & 30.78 & 28.6 \\
& & \;\; + \textbf{Ours (1$_{st}$)} & WRN-34-10 & \textbf{63.98} & \textbf{35.36} & \textbf{31.63} & \textbf{29.8} \\
& & \;\; + \textbf{Ours (1$_{st}$+2$_{nd}$)} & WRN-34-10 & \textbf{64.71} & \textbf{35.73} & \textbf{31.41} & \textbf{30.2} \\
\specialrule{.15em}{.075em}{.075em} 
\end{tabular}
}
\vspace{-3mm}
\end{table*}

\textbf{Evaluation Setting.} 
We evaluate the robustness with wihte-box attacks, black-box attacks and auto attack. For \textbf{white-box attacks}, we adopt 
PGD-20 \cite{DBLP:conf/iclr/MadryMSTV18} and CW-20~\cite{carlini2017towards} (the $\ell_\infty$ version of CW loss optimized by PGD-20) to evaluate trained models.
For \textbf{black-box attacks}, we generate adversarial perturbations by attacking a surrogate normal adversarial training model (with same setting)~\cite{papernot2017practical}, and then apply these adversarial examples to the defense model and evaluate the performances. 
The attacking methods for black-box attacks we have used are PGD-20 and CW-20.
For \textbf{Auto Attack (AA)}~\cite{croce2020reliable}, one of the strongest attack methods, we adopt it through a mixture of different parameter-free attacks which include three white-box attacks (APGD-CE~\cite{croce2020reliable}, APGD-DLR~\cite{croce2020reliable}, and FAB~\cite{croce2020minimally}) and one black-box attack (Square Attack~\cite{andriushchenko2020square}). We provide more details about the experimental setups in \cref{appendix:b}.

\subsection{Sensitivity of hyper-parameter}
\label{subsec:5.1}
In our proposed algorithm, the regularization hyper-parameter $\eta$ is crucial.
We use numerical experiments on CIFAR-10 with PreAct ResNet-18 to illustrate how the regularization hyper-parameter influences the performance of our robust classifiers.
To develop robust classifiers for multi-class tasks, we use the gradient descent update formula in Algorithm~\ref{alg}, with $\Lc$ as the cross-entropy loss.
Note that, for the zeroth term hyper-parameter $1/\lambda$ in Algorithm~\ref{alg}, we set $1/\lambda=6$ for all of our experiments, which is a widely used setting for TRADES. We train the models with $\eta=$ $0.05$, $0.1$, $0.2$, $0.3$, $0.4$ respectively. All models are trained with $200$ epochs and $128$ batchsize.

In Tab.~\ref{tab:1}, we observe that as the regularization parameter $\eta$ increases, the clean accuracy and robust accuracy almost both go up first and then  down. In particular, the accuracy under Auto Attack is sensitive to the regularization hyper-parameter $\eta$. Considering both robustness accuracy and clean accuracy, it is not difficult to find that $\eta=0.3$ is the best for first term optimization and $\eta=0.2$ is the best for first + second terms optimization. Thus in the following experiments, we set $\eta=0.3$ for first term optimization and $\eta=0.2$ for first + second terms optimization as default.

\begin{table*}[t]
\centering
\caption{Adversarial training across datasets on PreAct ResNet18 with $\ell_2$ threat model. Classification accuracy (\%) on clean images and under PGD-20 attack and Auto Attack. The results of our methods are in \textbf{bold}.}
\label{tab:3}
\vspace{-3mm}
\scalebox{1}{
\begin{tabular}{lccccccccccccc}
\specialrule{.15em}{.075em}{.075em} 
\multicolumn{1}{c}{\multirow{2}{*}{Method}} &&& \multicolumn{3}{c}{CIFAR-10} && \multicolumn{3}{c}{CIFAR-100} && \multicolumn{3}{c}{SVHN} \\ \cline{4-6} \cline{8-10} \cline{12-14}
&&& Clean & PGD & AA & & Clean & PGD & AA & & Clean & PGD & AA         
\\ \cline{0-1} \cline{4-6} \cline{8-10} \cline{12-14}
Wu et al. (2020) \cite{DBLP:conf/nips/WuX020}(0$_{th}$) &&& 87.05 & 72.08 & 71.6 && 62.87 & 45.11 & 41.8 && 92.86 & 72.45 & 63.8  \\
\;\; + \textbf{Ours (1$_{st}$+2$_{nd}$)} &&& \textbf{88.41} & \textbf{72.85} & \textbf{72.0} && \textbf{65.59} & \textbf{46.88} & \textbf{42.4} && \textbf{93.98} & \textbf{73.07} & \textbf{64.5}  \\ 
\specialrule{.15em}{.075em}{.075em} 
\end{tabular}
}
\vspace{-3mm}
\end{table*}
\begin{table*}[t]
\centering
\caption{Adversarial training across datasets on PreAct ResNet18 with $\ell_\infty$ threat model. Classification accuracy (\%) on clean images and black-box attacks. Black-box adversarial examples are generated by a surrogate normal adversarial training model (of same setting) with PGD-20 attack and CW-20 attack. The results of our methods are in \textbf{bold}.}
\label{tab:4}
\vspace{-3mm}
\scalebox{1}{
\begin{tabular}{lccccccccccccc}
\specialrule{.15em}{.075em}{.075em} 
\multicolumn{1}{c}{\multirow{2}{*}{Method}} &&& \multicolumn{3}{c}{CIFAR-10} && \multicolumn{3}{c}{CIFAR-100} && \multicolumn{3}{c}{SVHN} \\
\cline{4-6} \cline{8-10} \cline{12-14}
&&& Clean & PGD & CW & & Clean & PGD & CW & & Clean & PGD & CW         
\\ \cline{0-1} \cline{4-6} \cline{8-10} \cline{12-14}
Wu et al. (2020) \cite{DBLP:conf/nips/WuX020}(0$_{th}$) &&& 82.78 & 61.79 & 59.42 && 58.33 & 38.55 & 36.70 && 93.77 & 63.80 & 59.54   \\
\;\; + \textbf{Ours (1$_{st}$+2$_{nd}$)} &&& \textbf{84.27} & \textbf{62.56} & \textbf{60.58} && \textbf{61.08} & \textbf{39.82} & \textbf{37.84} && \textbf{95.11} & \textbf{66.16} & \textbf{63.59} \\ 
\specialrule{.15em}{.075em}{.075em} 
\end{tabular}
}
\vspace{1mm}
\centering
\caption{Adversarial training across VGG16, MobileNetV2 on CIFAR-10 with $\ell_\infty$ threat model. Classification accuracy (\%) on clean images and under PGD-20 attack, CW-20 attack and Auto Attack. The results of our methods are in \textbf{bold}.}
\label{tab:5}
\vspace{-2mm}
\scalebox{1}{
\begin{tabular}{lccccccccccc}
\specialrule{.15em}{.075em}{.075em} 
\multicolumn{1}{c}{\multirow{2}{*}{Method}} &&& \multicolumn{4}{c}{VGG16} && \multicolumn{4}{c}{MobileNetV2} \\
\cline{4-7} \cline{9-12}
&&& Clean & PGD-20 & CW-20 & AA & & Clean & PGD-20 & CW-20 & AA         
\\ \cline{0-1} \cline{4-7} \cline{9-12}
Zhang et al. (2019) \cite{zhang2019theoretically} (0$_{th}$) &&& 79.78 & 49.88 & 46.95 & 44.3 && 79.73 & 51.41 & 48.43 & 46.4  \\
\;\; + \textbf{Ours (1$_{st}$+2$_{nd}$)} &&& \textbf{80.99} & \textbf{50.13} & \textbf{47.09} & \textbf{44.4} && \textbf{81.86} & \textbf{53.34} & \textbf{49.93} & \textbf{47.8}   \\ 
\cline{0-1} \cline{4-7} \cline{9-12}
Wu et al. (2020) \cite{DBLP:conf/nips/WuX020} (0$_{th}$) &&& 78.46 & 51.19 & 47.41 & 46.3 && 79.86 & 53.56 & 50.11 & 47.7  \\
\;\; + \textbf{Ours (1$_{st}$+2$_{nd}$)} &&& \textbf{80.31} & \textbf{52.71} & \textbf{48.38} & \textbf{46.5} && \textbf{81.95} & \textbf{55.37} & \textbf{51.55} & \textbf{49.4} \\ 
\specialrule{.15em}{.075em}{.075em} 
\end{tabular}
}
\vspace{-4mm}
\end{table*}

\begin{table}[t]
\centering
\caption{Time consumption and GPU memory usage for WideResNet on CIFAR-10 with $\ell_\infty$ threat model.
We deploy each model on a single NVIDIA A100 with batchsize 128. The results of our methods are in \textbf{bold}.}
\label{tab:6}
\vspace{-3mm}
\scalebox{0.82}{
\begin{tabular}{lcccc}
\specialrule{.15em}{.075em}{.075em} 
\multicolumn{1}{c}{\multirow{2}{*}{Method}} &&& \multicolumn{2}{c}{WideResNet-34-10} \\
\cline{4-5} 
&&& Time/Epoch & GPU Memory        \\ 
 \cline{0-1} \cline{4-5} 
Zhang et al. (2019) \cite{zhang2019theoretically} (0$_{th}$) &&& 1666s & 12875MB  \\
\;\; + \textbf{Ours (1$_{st}$)}  &&& \textbf{2161s} & \textbf{20629MB}   \\ 
\;\; + \textbf{Ours (1$_{st}$+2$_{nd}$)} &&& \textbf{2362s} & \textbf{26409MB}   \\ 
\specialrule{.15em}{.075em}{.075em} 
\end{tabular}
}
\vspace{-4mm}
\end{table}

\subsection{Comparison with SOTA on WideResNet}
\label{subsec:5.2}

In \cref{tab:2}, we compare our method with state-of-the-art on WideResNet with CIFAR-10 and CIFAR-100. For zeroth term optimization, we use TRADES~\cite{zhang2019theoretically} and AWP-TRADES~\cite{DBLP:conf/nips/WuX020} as baselines with $1/\lambda=6$.
We also report other state-of-the-art, include AVMixup~\cite{DBLP:conf/cvpr/LeeLY20}, MART~\cite{DBLP:conf/iclr/0001ZY0MG20}, \cite{DBLP:conf/icml/RiceWK20}, \cite{DBLP:conf/icml/ZhangXH0CSK20}, \cite{DBLP:conf/iclr/PangYDSZ21}, TRADES+LBGAT~\cite{DBLP:conf/iccv/Cui0WJ21} and \cite{DBLP:journals/corr/abs-2010-03593}.

The results in Tab.~\ref{tab:2} demonstrate that our method can both improve clean accuracy and robustness accuracy consistently over different datasets and models. Especially for Auto Attack, our method can get $57.4\%$ on CIFAR-10 and $30.2\%$ on CIFAR-100 with WRN-34-10, even surpass the performance of WRN-70-16 from \cite{DBLP:journals/corr/abs-2010-03593}.

\subsection{Other empirical results}

\textbf{Adversarial training with $\ell_2$ threat model.} For the experiments with $\ell_2$ threat model on CIFAR-10, CIFAR-100 and SVHN in Tab.~\ref{tab:3}, the results still support that our method can enhance the performance under clean data, PGD-20 attack and Auto Attack. For CIFAR-100, it can even increase about $3\%$ on clean accuracy.

\textbf{Robustness under black-box attacks.} We train PreAct ResNet18 for $\ell_\infty$ threat model on CIFAR-10, CIFAR-100 and SVHN. The black-box adversarial examples are generated by a surrogate normal adversarial training model (of same setting) with PGD-20 attack and CW-20 attack. Tab.~\ref{tab:4} shows our method is also effective under black-box attacks. For SVHN, ours can even obtain a significant improvement over the existing ones.

\textbf{Robustness on other architectures.} We train VGG16 and MobileNetV2 for $\ell_\infty$ threat model on CIFAR-10. Our results in Tab.~\ref{tab:5} show a comprehensive improvement under clean data and robustness accuracy. Particularly, for MobileNetV2, ours can achieve an improvement greater than $1\%$ and $2\%$ under Auto Attack and clean data, respectively.

\textbf{Comparison with SWA from \cite{chen2020robust}.} 
The best PGD-20 accuracy and AA accuracy on CIFAR-10 for ResNet-18 from \cite{chen2020robust} are $52.14\%$ and $49.44\%$ respectively,
whereas ours are $55.13\%$ and $50.8\%$ in \cref{tab:1} (with 1$_{st}$+2$_{nd}$ (0.2)).

More empirical results are given in \Cref{appendix:more}.


\subsection{Limitations and future work}
We use some approximation methods, e.g., \cref{eq:upper}, to reduce the complexity of our adversarial training method.
The results in \cref{tab:6} show that though our method with WideResNet-34-10 can work on a single NVIDIA A100, the growth of training time and GPU memory still cannot be ignored.
In the future work, we plan to further reduce the complexity of our algorithm and apply the first and second terms optimization on larger datasets. 

\section{Related work}

\subsection{Adversarial training}

Adversarial training methods can generally be divided into three groups. In the first group, \cref{eq:15} is translated into an equivalent, (or approximate, expression), which is mainly used to narrow the distance between $f_\w(\s)$ and $f_\w(\s')$.
For example, ALP~\cite{engstrom2018evaluating,kannan2018adversarial} estimates the similarity between $f_\w(\s)$ and $f_\w(\s')$, and maximizes this similarity in the objective function.
MMA~\cite{ding2018mma} proposes each correctly classified instance $\s$ to leave sufficiently the decision boundary, through making the size of shortest successful perturbation as large as possible. 
TRADES~\cite{zhang2019theoretically} looks for the adversarial example $\s'$ which can obtain the largest KL divergence between $f_\w(\s)$ and $f_\w(\s')$, then trains with these adversarial examples.
Besides, \cite{Zheng_Chen_Ren_2019} adopts the similarity between local distributions of natural and adversarial example, \cite{SJ2017} measures the similarity of local distributions with Wasserstein distance, and  \cite{DBLP:conf/nips/MaoZYVR19,NEURIPS2020_5de8a360,DBLP:conf/cvpr/DongFYPSXZ20,DBLP:conf/iclr/PangYDSZ21} try to optimize with the distribution over a series of adversarial examples for a single input.

In the second group, adversarial examples are pre-processed before being used for training rather than being directly generated by attack algorithms.
For example, label smoothing~\cite{szegedy2016rethinking,chen2020robust} replaces the hard label $y$ with the soft label $\tilde{y}$, where $\tilde{y}$ is a combination of the hard label $y$ and uniform distribution.
A further empirical exploration with of how label smoothing works is provided in \cite{muller2019does}.
AVMixup~\cite{zhang2020does,DBLP:conf/cvpr/LeeLY20} defines a virtual sample in the adversarial direction based on these. Through linear interpolation of the virtual sample and the clean sample, it extends the training distribution with soft labels.
Instead of smoothing labels, \cite{NEURIPS2019_d8700cbd} perturbs the local neighborhoods with an unsupervised way to produce adversarial examples.
In addition, data augmentation~\cite{DBLP:journals/corr/abs-2103-01946,DBLP:conf/nips/GowalRWSCM21} is also shown to be an invaluable aid for adversarial training.

The above two groups merely adjust the component parts of the min-max formalism.
With AWP \cite{DBLP:conf/nips/WuX020}, one more maximization is performed to find a weight perturbation based on the generated adversarial examples. 
The altered weights~\cite{devries2017improved} are then used in the outer minimization function to reduce the loss caused by the adversarial cases.

Different from previous work, this paper optimizes the distance between $f_{\w+\uu}(\s)$ and $f_{\w+\uu}(\s')$ through randomized weights, and therefore, effectively, it considers the optimization problem where the decision boundary is smoothed by these randomized models.


\subsection{Randomized weights}
In the previous work to study generalization or robustness, modeling neural network weights as random variables has been frequently used. For example, \cite{xu2017information,negrea2019information} estimate the mutual information between random weights and dataset, provide the expected generalization bound over weight distribution.
Another example of random weights is on the PAC-Bayesian framework, a well-known theoretical tool to bound the generalization error of machine learning models~\cite{DBLP:conf/colt/McAllester99,DBLP:journals/jmlr/Seeger02,DBLP:conf/nips/LangfordS02,DBLP:journals/jmlr/Parrado-HernandezASS12}. 
In the recent years, PAC-Bayes is also 
developed to bound the generalization error or robustness error of deep neural networks~\cite{DBLP:conf/uai/DziugaiteR17,DBLP:conf/nips/NeyshaburBMS17,DBLP:conf/iclr/FarniaZT19,DBLP:conf/iclr/JiangNMKB20}.
In addition, \cite{DBLP:conf/icml/GoldfeldBGMNKP19} draws random noise into deterministic weights, to measure the mutual information between input dataset and activations under information bottleneck theory.

While we also consider randomized weights, a major difference from previous theoretical works is that randomized weights are applied to robustness, from which we find an insight to design a novel adversarial training method which decomposes objective function (parameterized with randomized weights) with Taylor series.

\subsection{Flatness}

The generalization performance of machine learning models is believed to be correlated with the flatness of loss curve \cite{DBLP:conf/nips/HochreiterS94,DBLP:journals/neco/HochreiterS97a}, especially for  deep neural networks \cite{DBLP:conf/icml/DinhPBB17,DBLP:conf/iclr/ChaudhariCSLBBC17,DBLP:conf/nips/YaoGLKM18,DBLP:conf/aistats/LiangPRS19,DBLP:conf/nips/JinYZ0S020,mu2022certified}.
For instance, in pioneering research \cite{DBLP:conf/nips/HochreiterS94,DBLP:journals/neco/HochreiterS97a}, the minimum description length (MDL) \cite{DBLP:journals/automatica/Rissanen78}  was used to demonstrate that the flatness of the minimum is a reasonable generalization metric to consider.
\cite{DBLP:conf/iclr/KeskarMNST17} investigated the cause of the decrease in generalization for large-batch networks and demonstrated that large-batch algorithms tend to converge to sharp minima, resulting in worse generalization.
\cite{DBLP:conf/iclr/JiangNMKB20} undertook a large-scale empirical study and found that flatness-based measurements correlate with generalization more strongly than weight norms and (margin and optimization)-based measures.
\cite{DBLP:conf/iclr/ForetKMN21} introduced Sharpness-Aware Minimization (SAM) to locate parameters that lie in neighborhoods with consistently low loss. 
Moreover, 
\cite{chen2020robust} 
used SWA \cite{DBLP:conf/uai/IzmailovPGVW18} in adversarial training to smooth the weights and find flatter minima, and showed that this can improve adversarial generalization.

Different from previous work, this paper uses Taylor expansion of the injected Gaussian to smooth the update of weights and search for flat minima during adversarial training.
Our empirical results show the effectiveness of our method to flatten loss landscape and improve both robustness and clean accuracy.

\section{Conclusion}

This work studies the trade-off between robustness and clean accuracy through the lens of randomized weights. 
Through empirical analysis of loss landscape, 
algorithmic design (via Taylor expansion) on training optimization, and extensive experiments, we demonstrate that optimizing over randomized weights can consistently outperform the state-of-the-art adversarial training methods not only in clean accuracy but also in robustness. 

\vspace{-3mm}
\paragraph{Acknowledgment.} 
GJ is supported by the CAS Project for Young Scientists in Basic Research, Grant No.YSBR-040. 
This project has received funding from the European Union’s Horizon 2020 research and innovation programme under grant agreement No 956123, and is also supported by the U.K. EPSRC through End-to-End Conceptual Guarding of Neural Architectures [EP/T026995/1].

\newpage
{\small
\bibliographystyle{ieee_fullname}
\bibliography{egbib}
}

\clearpage

\appendix
\section{More results of \cref{fig:2}}
\label{appendix:a}

\begin{figure*}[t!]
\includegraphics[width=1
\textwidth]{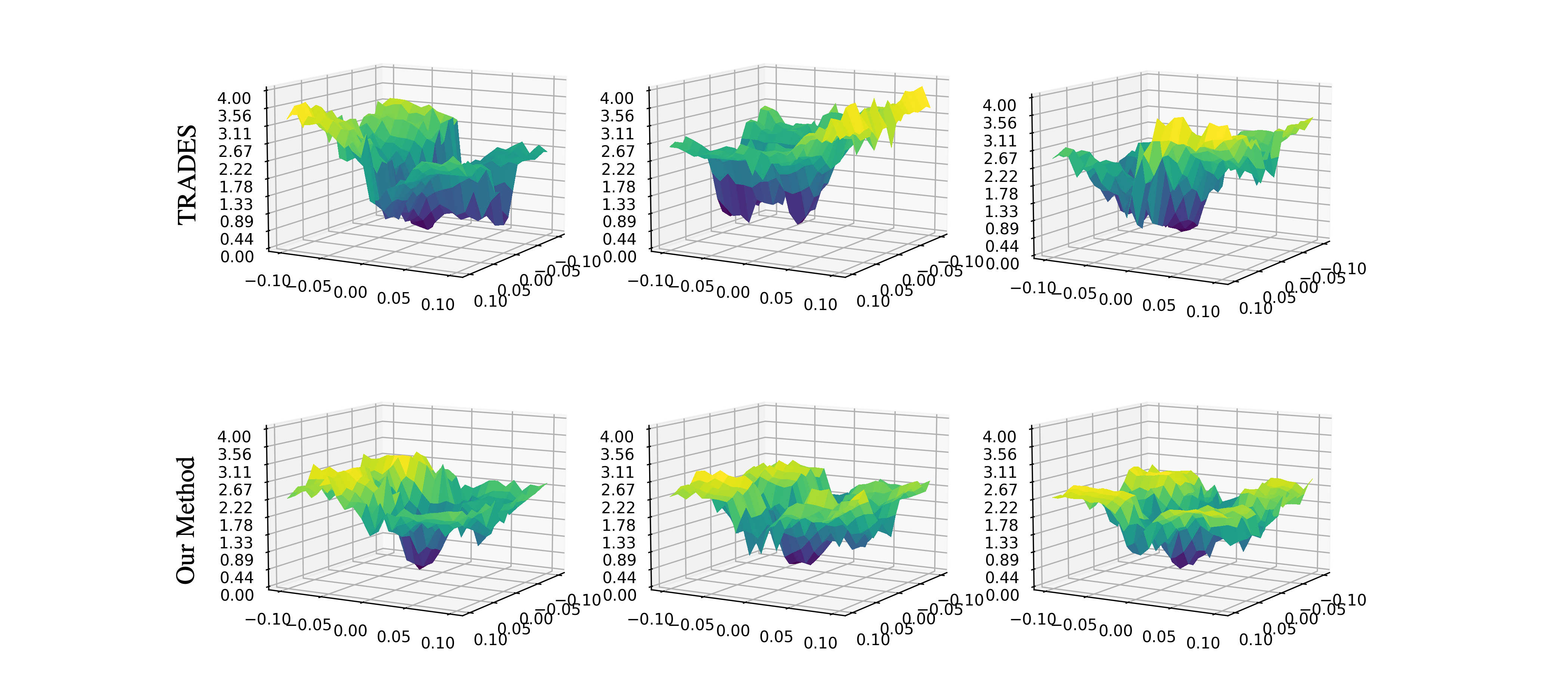}
\centering
\caption{Comparison of loss landscapes of TRADES trained model (the first row) and TRADES+$1_{st}$+$2_{nd}$ (our method, the second row) trained model.
Loss plots in each column are generated from the same original image randomly chosen from the CIFAR-10 test dataset.
Following the settings in \cite{DBLP:journals/corr/abs-1807-10272}, the $z$ axis represents the loss, the $x$ and $y$ axes represent the magnitude of the perturbation added in the directions of $\vec{sign}\nabla_{\s}f(\s)$ and Rademacher($0.5$) respectively.}
\label{fig:3}
\end{figure*}

In \cref{fig:3}, we provide additional empirical results of CIFAR-10 on ResNet-18, which show that our method can effectively flatten the loss landscape and find flat minima.

\section{Details of the experiments}
\label{appendix:b}

\subsection{Network Architecture}

For all of our experiments in Sec.~\ref{sec:experiment}, we use 4 network architectures as ResNet, WideResNet, VGG and MobileNetV2. 
We present the details in the following.

\begin{itemize}

\item ResNet/WideResNet: Architectures used are PreAct ResNet. All convolutional layers (except downsampling convolutional layers) have kernel size $3\times 3$ with stride $1$. Downsampling convolutions have stride 2. All the ResNets have five stages (0-4) where each stage has multiple residual/downsampling blocks. These stages are followed by a max-pooling layer and a
final linear layer. We study the PreAct ResNet 18 and WideResNet-34-10. 

\item VGG: Architecture consists of multiple convolutional layers, followed by multiple fully connected layers and a final classifier layer (with output dimension 10 or 100). We study the VGG networks with 16 layers.

\item MobileNetV2: Architecture is built on an inverted residual structure, with residual connections between bottleneck layers. As a source of non-linearity, the intermediate expansion layer filters features with lightweight depthwise convolutions. As a whole, the architecture of MobileNetV2 includes a fully convolutional layer with three filters, followed by 19 residual bottleneck layers.
\end{itemize}

\subsection{Checkpoints}
We set checkpoints for each epoch between $100-110$ and $150-160$, each 5 epoch between $110-150$ and $160-200$. All best performances are gotten from these checkpoints.

\section{An un-rigorous theoretical perspective}
\label{appendix:c}
\emph{Please note that we provide a \textbf{potential, interesting but un-rigorous} theoretical perspective in the following. This section is \textbf{not claimed as the contribution} of this paper.}

This section explores theoretical implication of the use of randomized weights on robustness. 
Specifically, we provide a \textbf{shallow} theoretical perspective which discusses how randomized weights may affect the information-theoretic generalization bound of both clean and adversarial data.

Under the information-theoretic context, a learning algorithm can be taken as a randomized mapping, where training data set is input and hypothesis is output.
With that, \cite{xu2017information} considered a generalization bound based on the information contained in weights $I(\Lambda_\WW(\St);\w)$, where
$\Lambda_\WW(\St):= \big(\Lc \big(f_\w(\St), \mathcal{Y} \big)\big)_{\w \in \WW}$
is the collection of empirical losses in hypotheses space $\WW$.
Let $\PP(\Lc (f_{\w_1}(\St), \mathcal{Y} ), \w_2)=0$ where $\w_1 \neq \w_2$, 
we can use $I(\Lc(f_{\ww}(\St), \mathcal{Y});\w)$ to approximate $I(\Lambda_\WW(\St);\w)$ where $\w$ is the randomized weight distributed in $\WW$, then get the following perspective.

Suppose $\Lc(f_{\w}(\s), y)$ is $\sigma_*$-sub-Gaussian, $\D$ is the clean data distribution and $\St$ is the training data set with $m$ samples, then 
\begin{equation}
\begin{aligned}
    &\quad\;\E_{\w}\Big( \Lc \big(f_\w(\D), \mathcal{Y} \big)-\Lc \big(f_\w(\St),\mathcal{Y} \big) \Big)\\
    &\le \sqrt{\frac{2\sigma_*^2}{m}I\Big(\Lc \big(f_{\ww}(\St),\mathcal{Y} \big);\w|\St\Big)}.
\end{aligned}
\end{equation}

The above inequation provides the upper bound on the expected generalization error of randomized weights.
Building upon the above bound, we consider generalization errors of both clean and adversarial data based on discrete distribution, then obtain the following proposition.

\begin{myprop}
\label{thm:c.1}
Let $\Lc(f_{\w+\U}(\s), y)$ and $\Lc(f_{\w+\U}(\s'), y)$ be $\sigma_*$-sub-Gaussian.
We suppose the adversarial trained model contains information of $\St\vee \St'$, where $\St\vee\St'$ is the joint set and 
the training samples are chosen at random from $\St$ and $\St'$ with probabilities $q$ and $1-q$,
i.e., $\Lc(f_{\w+\U}(\St\vee\St'), \mathcal{Y})=q\Lc(f_{\w+\U}(\St), \mathcal{Y})+(1-q)\Lc(f_{\w+\U}(\St'), \mathcal{Y})$.
We let $q \in (0, 0.5)$, since $\St'$ occupies a large proportion in adversarial training, then
\begin{small}
\begin{equation}
\begin{aligned}
    &\quad\;\E_{\w+\U}\Big( \Lc \big(f_{\w+\U}(\D\vee\D'), \mathcal{Y} \big)-\Lc \big(f_{\w+\U}(\St\vee\St'),\mathcal{Y} \big) \Big)\\
    &\le \sqrt{\frac{2\sigma_*^2}{m}I\Big(\Lc \big(f_{\ww+\uu}(\St\vee\St'),\mathcal{Y} \big);\w+\U|\St\vee\St'\Big)}\\
    &\le \sqrt{\frac{2\sigma_*^2}{m}I\Big(\Lc \big(f_{\ww+\uu}(\St),\mathcal{Y} \big),\Lc \big(f_{\ww+\uu}(\St'),\mathcal{Y} \big);\w+\U|\St\vee\St'\Big)}.
\end{aligned}
\end{equation}
\end{small}
\end{myprop}

We now make several observations about above inequation.
First, it is obvious that more training data (larger $m$) helps adversarial training to get a high-performance model.
Second, take into account both generalization errors of clean and adversarial data with coefficient $q$, a lower mutual information between $\big(\Lc \big(f_{\ww+\uu}(\St),\mathcal{Y} \big),\Lc \big(f_{\ww+\uu}(\St'),\mathcal{Y} \big)\big)$ and $\w+\U$ is essential to get a better performance of robustness and clean accuracy. 

The above mutual information is a statistic over high-dimensional space, thus we are almost impossible to directly estimate and optimize it during training. 
Nevertheless, we can reduce it implicitly through the following Lemma.

\begin{mylem}
\label{thm:c.2}
$I\big(\Lc \big(f_{\ww+\uu}(\St),\mathcal{Y} \big),\Lc \big(f_{\ww+\uu}(\St'),\mathcal{Y} \big);\w+\U\big)$ is lower bounded by $I\big(\Lc \big(f_{\ww+\uu}(\St),\mathcal{Y} \big);$ $\w+\U\big)$ and upper bounded by $I\big(\Lc \big(f_{\ww+\uu}(\St),\mathcal{Y} \big);\w+\U\big)+I\big(\Lc \big(f_{\ww+\uu}(\St'),\mathcal{Y} \big);\w+\U\big)$, i.e.,
\begin{small}
\begin{equation}
\begin{aligned}
    &I\big(\Lc \big(f_{\ww+\uu}(\St),\mathcal{Y} \big);\w+\U\big)\\ 
    &\le I\big(\Lc \big(f_{\ww+\uu}(\St),\mathcal{Y} \big),\Lc \big(f_{\ww+\uu}(\St'),\mathcal{Y} \big);\w+\U\big)\\ 
    &\le I\big(\Lc \big(f_{\ww+\uu}(\St),\mathcal{Y} \big);\w+\U\big)+I\big(\Lc \big(f_{\ww+\uu}(\St'),\mathcal{Y} \big);\w+\U\big).
\end{aligned}
\end{equation}
\end{small}
\end{mylem}

\begin{custompro}{C.2}
We suppose the adversarial trained model contains information of $\St\vee \St'$, where $\St\vee\St'$ is the joint set with $\Lc(f_{\w+\U}(\St\vee\St'), \mathcal{Y})=q\Lc(f_{\w+\U}(\St), \mathcal{Y})+(1-q)\Lc(f_{\w+\U}(\St'), \mathcal{Y})$.
We let $q \in (0, 0.5)$ because $\St'$ occupies a large proportion in adversarial training.
Randomized weight $\w$ is generated by normal training with data set $\St$ and $\w+\U$ is generated by adversarial training with joint set $\St\vee \St'$, then
\begin{small}
\begin{equation}\nonumber
\begin{aligned}
    &I\Big(\Lc \big(f_{\ww+\uu}(\St),\mathcal{Y} \big),\Lc \big(f_{\ww+\uu}(\St'),\mathcal{Y} \big);\w+\U\Big)\\
    &=H\Big(\Lc \big(f_{\ww+\uu}(\St),\mathcal{Y} \big),\Lc \big(f_{\ww+\uu}(\St'),\mathcal{Y} \big)\Big)\\
    &\quad-H\Big(\Lc \big(f_{\ww+\uu}(\St),\mathcal{Y} \big),\Lc \big(f_{\ww+\uu}(\St'),\mathcal{Y} \big)\Big|\w+\U\Big)\\
    &=H\Big(\Lc \big(f_{\ww+\uu}(\St),\mathcal{Y} \big)\Big)+H\Big(\Lc \big(f_{\ww+\uu}(\St'),\mathcal{Y} \big)\Big|\Lc \big(f_{\ww+\uu}(\St),\mathcal{Y} \big)\Big)\\
    &\quad-H\Big(\Lc \big(f_{\ww+\uu}(\St),\mathcal{Y} \big)\Big|\w+\U \Big)\\
    &\quad-H\Big(\Lc \big(f_{\ww+\uu}(\St'),\mathcal{Y} \big)\Big|\Lc \big(f_{\ww+\uu}(\St),\mathcal{Y} \big),\w+\U\Big)\\
    &=H\Big(\Lc \big(f_{\ww+\uu}(\St),\mathcal{Y} \big)\Big)+H\Big(\Lc \big(f_{\ww+\uu}(\St'),\mathcal{Y} \big)\Big)\\
    &\quad-I\Big(\Lc \big(f_{\ww+\uu}(\St),\mathcal{Y} \big);\Lc \big(f_{\ww+\uu}(\St'),\mathcal{Y} \big)\Big)\\
    &\quad-H\Big(\Lc \big(f_{\ww+\uu}(\St),\mathcal{Y} \big)\Big|\w+\U\Big)-H\Big(\Lc \big(f_{\ww+\uu}(\St'),\mathcal{Y} \big)\Big|\w+\U\Big)\\
\end{aligned}    
\end{equation}
\end{small}
\begin{small}
\begin{equation}
\begin{aligned}
    &\quad+I\Big(\Lc \big(f_{\ww+\uu}(\St),\mathcal{Y} \big);\Lc \big(f_{\ww+\uu}(\St'),\mathcal{Y} \big)\Big|\w+\U\Big)\\
    &= I\Big(\Lc \big(f_{\ww+\uu}(\St),\mathcal{Y} \big);\w+\U\Big)+I\Big(\Lc \big(f_{\ww+\uu}(\St'),\mathcal{Y} \big);\w+\U\Big)\\
    &\quad-I\Big(\Lc \big(f_{\ww+\uu}(\St),\mathcal{Y} \big);\Lc \big(f_{\ww+\uu}(\St'),\mathcal{Y} \big);\w+\U\Big).
\end{aligned}    
\end{equation}
\end{small}Note that $I\big(\Lc \big(f_{\ww+\uu}(\St),\mathcal{Y} \big);\w+\U\big)\le I\big(\Lc \big(f_{\ww+\uu}(\St'),\mathcal{Y} $ $\big);\w+\U\big)$ holds, due to the data processing inequality and $\St-\St\vee\St'-\w+\ul$ forms a Markov chain under adversarial training where $\St'$ occupies a larger proportion in $\St\vee\St'$. As $I\big(\Lc \big(f_{\ww+\uu}(\St'),\mathcal{Y} \big);\w+\U\big)\ge I\big(\Lc \big(f_{\ww+\uu}(\St),\mathcal{Y} \big);\Lc \big(f_{\ww+\uu}(\St'),\mathcal{Y} \big);\w+\U\big)$, we can easily get Lem.~\ref{thm:c.2}. 
\hfill $\square$
\end{custompro}
Lem.~\ref{thm:c.2} gives us an upper bound and a lower bound.
The upper bound represents the worst case of adversarial trained model where $\Lc \big(f_{\ww+\uu}(\St),\mathcal{Y} \big)$ and $\Lc \big(f_{\ww+\uu}(\St'),\mathcal{Y} \big)$ are radically different (uncorrelated).
To some extent, it means the trained model is failed to extract common features of clean data and adversarial data, thus needs to use more parameters to recognize clean and adversarial examples respectively.
Lem.~\ref{thm:c.2} also charts a realizable optimization direction for the model, the lower bound $I\big(\Lc \big(f_{\ww+\uu}(\St),\mathcal{Y} \big);\w+\U\big)$ is the optimal case of adversarial training for $I\big(\Lc \big(f_{\ww+\uu}(\St),\mathcal{Y} \big),\Lc \big(f_{\ww+\uu}(\St'),\mathcal{Y} \big);\w+\U\big)$.
In this situation, $\Lc \big(f_{\ww+\uu}(\St),\mathcal{Y} \big)$ and $\Lc \big(f_{\ww+\uu}(\St'),\mathcal{Y} \big)$ are completely correlated,
that is, the optimal adversarial trained model is successful at extracting common features of clean data and adversarial data.

It is obvious that $I\big(\Lc \big(f_{\ww+\uu}(\St),\mathcal{Y} \big),\Lc \big(f_{\ww+\uu}(\St'),\mathcal{Y} \big);$ $\w+\U\big)$ is still difficult to be estimated in a high-dimensional space. Fortunately, Lem.~\ref{thm:c.2} allows us to optimize it via narrowing the distance between $\Lc \big(f_{\ww+\uu}(\St),\mathcal{Y} \big)$ and $\Lc \big(f_{\ww+\uu}(\St'),\mathcal{Y} \big)$, 
which makes $I\big(\Lc \big(f_{\ww+\uu}(\St),\mathcal{Y} \big),\Lc \big(f_{\ww+\uu}(\St'),\mathcal{Y} \big);\w+\U\big)$ close to the lower bound $I\big(\Lc \big(f_{\ww+\uu}(\St),\mathcal{Y} \big);\w+\U\big)$.

\begin{mylem}
\label{thm:c.3}    
In this lemma, we consider the case of binary response $y \in \{0, 1\}$, 
then the gap between $\Lc \big(f_{\ww+\uu}(\s),y \big)$ and $\Lc \big(f_{\ww+\uu}(\s'),y \big)$ is positive correlated with the KL divergence between $f_{\ww+\uu}(\s)$ and $f_{\ww+\uu}(\s')$, i.e.,
\begin{equation}
\begin{aligned}
    &|\Lc \big(f_{\ww+\uu}(\s),y \big)-\Lc \big(f_{\ww+\uu}(\s'),y \big)| \\
    &\propto \KL\big(f_{\ww+\uu}(\s)||f_{\ww+\uu}(\s')\big),
\end{aligned}
\end{equation}
where we let $\propto$ represent positive correlation.
\end{mylem}

\begin{custompro}{C.3}
Let $f_{\ww+\uu}(\s)_{true}$ be the normalized output of $f_{\ww+\uu}(\s)$ for true label and we consider the case of binary response $y \in \{0, 1\}$ in Lem.~\ref{thm:c.3}. Then,
\begin{equation}
\begin{aligned}
    &\quad |\Lc \big(f_{\ww+\uu}(\s),y \big)-\Lc \big(f_{\ww+\uu}(\s'),y \big)|\\
    &=\Big|\log \frac{f_{\ww+\uu}(\s)_{true}}{f_{\ww+\uu}(\s')_{true}}\Big|\\
    &\propto f_{\ww+\uu}(\s)_{true}\log \frac{f_{\ww+\uu}(\s)_{true}}{f_{\ww+\uu}(\s')_{true}}\\
    &\quad + (1-f_{\ww+\uu}(\s)_{true})\log \frac{1-f_{\ww+\uu}(\s)_{true}}{1-f_{\ww+\uu}(\s')_{true}}\\
    &=\KL\big(f_{\ww+\uu}(\s)||f_{\ww+\uu}(\s')\big),
\end{aligned}
\end{equation}
where we let $\propto$ represent positive correlation.
\hfill $\square$
\end{custompro}
Lem.~\ref{thm:c.3} demonstrates $|\Lc (f_{\ww+\uu}(\s),y )-\Lc (f_{\ww+\uu}(\s'),y )|$ is positive correlated with $\KL(f_{\ww+\uu}(\s)||f_{\ww+\uu}(\s'))$ in a binary case, this can also approximately hold in a multi-class case.
Thus, it allows us to optimize the mutual information utilizing a simplified term of $\KL (f_{\w+\uu}(\s)||f_{\w+\uu}(\s'))$.
Although we are still difficult to directly deal with this $\KL$ term during training, it can be decomposed by our method in Sec.~\ref{sec:method} with Taylor series.

\section{Optimization}
\label{appendix:d}

It is easy to see that minimizing $\E_{\mathbf{u}}(\Lc(g'_{\s}(\w)^T\mathbf{u},$ $g'_{\s'}(\w)^T\mathbf{u}))$, 
$\E_{\mathbf{u}}(\Lc(\mathbf{u}^Tg''_{\s}(\w)\mathbf{u},\mathbf{u}^Tg''_{\s'}(\w)\mathbf{u}))$ is equivalent to reducing the distance between $g'_{\s}(\w)$ and $g'_{\s'}(\w)$, $g''_{\s}(\w)$ and $g''_{\s'}(\w)$, respectively.
Normally, the $\ell_2$ distance between vectors $g'_{\s}(\w)$ and $g'_{\s'}(\w)$ can be defined as
\begin{equation}
\begin{aligned}
    \Big|\Big|\frac{\partial g_{\s}(\w)}{\partial \w}-\frac{\partial g_{\s'}(\w)}{\partial \w}\Big|\Big|_2^2
    =\sum_{j} \sum_i\Big[\frac{\partial (g_{\s}(\w)-g_{\s'}(\w))}{\partial \W_{(j,i)}}\Big]^2.
\end{aligned}
\label{eq:17}
\end{equation}
We extend Eq.~(\ref{eq:17}) by considering the sum of each row vector of $\frac{\partial (g_{\s}(\w)-g_{\s'}(\w))}{\partial \W}$ and define the distance between $\frac{\partial g_{\s}(\w)}{\partial \w}$ and $\frac{\partial g_{\s'}(\w)}{\partial \w}$ as
\begin{equation}
    \sum_{j} \Big[\sum_i\frac{\partial (g_{\s}(\w)-g_{\s'}(\w))}{\partial \W_{(j,i)}}\Big]^2.
\end{equation}
We notice that, according to chain rule, $\Big[\sum_i\frac{\partial (g_{\s}(\w)-g_{\s'}(\w))}{\partial \W_{(j,i)}}\Big]^2$ corresponds to $g_{\s}(\w)_j$ and $g_{\s'}(\w)_j$, 
where $g_{\s}(\w)_j$ and $g_{\s'}(\w)_j$ are the $j$-th output of $g_{\s}(\w)$ and $g_{\s'}(\w)$ respectively.
Thus we can easily increase $g_{\s'}(\w)_j$ to reduce $\sum_i\frac{\partial (g_{\s}(\w)-g_{\s'}(\w))}{\partial \W_{(j,i)}}$ when $\sum_i\frac{\partial (g_{\s}(\w)-g_{\s'}(\w))}{\partial \W_{(j,i)}}$ is large.
In contrast, we can increase $g_{\s}(\w)_j$ to reduce $\sum_i\frac{\partial (g_{\s'}(\w)-g_{\s}(\w))}{\partial \W_{(j,i)}}$ when $\sum_i\frac{\partial (g_{\s'}(\w)-g_{\s}(\w))}{\partial \W_{(j,i)}}$ is large.
Then, we can approximately optimize $\E_{\mathbf{u}}(\Lc(g'_{\s}(\w)^T\mathbf{u},  g'_{\s'}(\w)^T\mathbf{u}))$ through minimizing
\begin{small}
\begin{equation}
\begin{aligned}
    &\frac{1}{2}\Lc\Big(g_{\s'}(\w),\\
    &\quad \Big[\sum_i\frac{\partial (g_{\s}(\w)-g_{\s'}(\w))}{\partial \W_{(1,i)}},...,\sum_i\frac{\partial (g_{\s}(\w)-g_{\s'}(\w))}{\partial \W_{(N,i)}}\Big]^T\Big)\\
    &+\frac{1}{2}\Lc\Big(g_{\s}(\w),\\
    &\quad \Big[\sum_i\frac{\partial (g_{\s'}(\w)-g_{\s}(\w))}{\partial \W_{(1,i)}},...,\sum_i\frac{\partial (g_{\s'}(\w)-g_{\s}(\w))}{\partial \W_{(N,i)}}\Big]^T\Big),
\end{aligned}
\end{equation}
\end{small}where $\W_{(j,k)}$ is the element of $j$-th row, $k$-th column of weight matrix $\W$, $N$ is the number of neurons (units) on the layer.

Similarly, we define the distance between $\frac{\partial^2 g_{\s}(\w)}{\partial \w \partial \w}$ and $\frac{\partial^2 g_{\s'}(\w)}{\partial \w \partial \w}$ as
\begin{equation}
    \sum_{l=1}^{N} \Big[\sum_i\sum_j\sum_k\frac{\partial^2 g_{\s}(\w)-\partial^2 g_{\s'}(\w)}{\partial \W_{(l,i)}\partial \W_{(j,k)}}\Big]^2.
\end{equation}
We also notice that, according to chain rule, $\Big[\sum_i\sum_j\sum_k$ $\frac{\partial^2 g_{\s}(\w)-\partial^2 g_{\s'}(\w)}{\partial \W_{(l,i)}\partial \W_{(j,k)}}\Big]^2$ corresponds to $g_{\s}(\w)_l$ and $g_{\s'}(\w)_l$.
We can increase $g_{\s'}(\w)_l$ to reduce $\sum_i\sum_j\sum_k$ $\frac{\partial^2 g_{\s}(\w)-\partial^2 g_{\s'}(\w)}{\partial \W_{(l,i)}\partial \W_{(j,k)}}$ when 
$\sum_i\sum_j\sum_k\frac{\partial^2 g_{\s}(\w)-\partial^2 g_{\s'}(\w)}{\partial \W_{(l,i)}\partial \W_{(j,k)}}$ is large.
In contrast, we can increase $g_{\s}(\w)_l$ to reduce $\sum_i\sum_j$ $\sum_k\frac{\partial^2 g_{\s'}(\w)-\partial^2 g_{\s}(\w)}{\partial \W_{(l,i)}\partial \W_{(j,k)}}$ when $\sum_i\sum_j\sum_k\frac{\partial^2 g_{\s'}(\w)-\partial^2 g_{\s}(\w)}{\partial \W_{(l,i)}\partial \W_{(j,k)}}$ is large.
Thus, we can approximately optimize $\E_{\mathbf{u}}(\Lc(\mathbf{u}^Tg''_{\s}(\w)\mathbf{u},\mathbf{u}^Tg''_{\s'}(\w)\mathbf{u}))$ through minimizing
\begin{equation}
\begin{aligned}
    &\frac{1}{2}\Lc\Big(g_{\s'}(\w),\Big[\sum_i\sum_j\sum_k\frac{\partial^2 g_{\s}(\w)-\partial^2 g_{\s'}(\w)}{\partial \W_{(1,i)}\partial \W_{(j,k)}},...,\\
    &\sum_i\sum_j\sum_k\frac{\partial^2 g_{\s}(\w)-\partial^2 g_{\s'}(\w)}{\partial \W_{(N,i)}\partial \W_{(j,k)}}\Big]^T\Big)\\
    &+\frac{1}{2}\Lc\Big(g_{\s}(\w),\Big[\sum_i\sum_j\sum_k\frac{\partial^2 g_{\s'}(\w)-\partial^2g_{\s}(\w)}{\partial \W_{(1,i)}\partial \W_{(j,k)}},...,\\
    &\sum_i\sum_j\sum_k\frac{\partial^2 g_{\s'}(\w)-\partial^2 g_{\s}(\w)}{\partial \W_{(N,i)}\partial \W_{(j,k)}}\Big]^T\Big).
\end{aligned}
\end{equation}

\section{More empirical results}
\label{appendix:more}

\begin{table}[t!]
\centering
\caption{CIFAR-10, PreAct ResNet 18, under RayS hard label attack ($\ell_\infty$,\%).}
\label{tab:7}
\vspace{-3mm}
\scalebox{0.7}{
\begin{tabular}{lccc}
\toprule[1.5pt]

Method & Clean Acc & ADBD & RayS Acc \\ \hline 

TRADES & 82.89 & 0.0412 & 56.06 \\
TRADES+1$_{st}$+2$_{nd}$  & 84.13 & 0.0435 & 57.03 \\

\toprule[1.5pt]
\end{tabular}
}
\vspace{-2mm}
\end{table}

\begin{table}[t!]
\centering
\caption{CIFAR-10, ResNet 18, comparison of our method on AT with GAT and HAT ($\ell_\infty$,\%).}
\label{tab:8}
\vspace{-3mm}
\scalebox{0.7}{
\begin{tabular}{lcccc}
\toprule[1.5pt]

Method & Clean & PGD-20 & CW-20 & AA \\ \hline 

AT & 82.41 & 52.77 & 50.43 & 47.1 \\
AT+1$_{st}$+2$_{nd}$ & 83.56 & 54.23 & 52.19 & 48.7 \\
GAT & 80.49 & 53.13 & - & 47.3  \\
TRADES+GAT & 81.32 & 53.37 & - & 49.6  \\
HAT & 84.90 & 49.08 & - & -  \\
HAT(DDPM) & 86.86 & 57.09 & - & -  \\

\toprule[1.5pt]
\end{tabular}
}
\vspace{-2mm}
\end{table}

More results of RayS hard-label attack \cite{chen2020rays}, HAT \cite{rade2022reducing
}, GAT \cite{sriramanan2020guided} methods are given in Tabs.~\ref{tab:7} and \ref{tab:8}.

\end{document}